# Predicting invasive ductal carcinoma using a Reinforcement Sample Learning Strategy using Deep Learning

Rushabh Patel[a]

[a]*Temple University, Philadelphia, PA. USA*

**Abstract**

Invasive ductal carcinoma is a prevalent, potentially deadly disease associated with a high rate of morbidity and mortality. Its malignancy is the second leading cause of death from cancer in women. The mammogram is an extremely useful resource for mass detection and invasive ductal carcinoma diagnosis. We are proposing a Computer Aided Diagnosis (CAD) method for Invasive ductal carcinoma that will use convolutional neural networks (CNN) on mammograms to assist radiologists in diagnosing the disease. Due to the varying image clarity and structure of certain mammograms, it is difficult to observe major cancer characteristics such as microcalcification and mass, and it is often difficult to interpret and diagnose these attributes. The aim of this study is to establish a novel method for fully automated feature extraction and classification in invasive ductal carcinoma computer-aided diagnosis (CAD) systems. This article presents a tumor classification algorithm that makes novel use of convolutional neural networks on breast mammogram images to increase feature extraction and training speed. The algorithm makes two contributions. To begin, a new CNN model was developed to automatically isolate and identify features from mammograms with broad image sizes. In contrast to other classification methods, it is fully automated and removes the need for manually identifying regions of interest, detecting tumors and microcalcifications, and extracting and describing characteristics. Second, to train the CNN with less iteration epochs and training time, a novel reinforcement sample learning scheme is proposed. To minimize training time, the proposed model is learned and evaluated using the Digital Database for Mammography Screening (DDSM). 1000 benign and malignant cases are randomly selected for training, and another 450 benign and malignant cases are used for validation. The experimental findings indicate that the CNN model trained with the latest training scheme will produce a 11 percent error within 150 epochs while the error in the validation range is less than 22 percent. It shows the viability of using CNN for large-scale feature extraction and classification in medical images, as well as the ability to use CNN for large-scale feature extraction and classification in medical images. Due to its quick training speed, the proposed algorithm will find widespread application in the image classification field, particularly due to the specific relationship between Convolutional Neural Networks and the used Reinforcement Sample Learning Strategy.

Keywords: Computer Aided Diagnosis System, Machine Learning, Computer Vision, Convolution Neural Networks, reinforcement sample learning, image classification

## 1. Introduction

Invasive ductal carcinoma is a prevalent condition with a high morbidity and mortality rate, and it is the second leading cause of death from cancer in women [1]. For more precise detection, computer-aided diagnosis (CAD) systems have been developed for detecting and classifying invasive ductal carcinoma. A CAD framework is an interdisciplinary technique that leverages both automated image processing algorithms and the expertise of radiologists to significantly increase the precision of abnormality identification and diagnosis [2]. A standard CAD method is composed of multiple steps, including image preprocessing, segmentation, feature extraction and reduction, and classification. Automated feature extraction and selection is a critical part of a computer-aided diagnostic (CAD) method for invasive ductal carcinoma. Features are extracted to classify the segmented tumor area and are then used for further classification. Traditionally, feature extraction and selection have relied on professional manual breast detection as a requirement for diagnosing possible diseases [3]. Classification of the breast is complicated due to its spatial complexity and heterogeneity. Even though considerable study has been conducted in this area and several feature descriptors have been proposed over the years, they may be very complex and domain specific. Selecting significant features for classification remains a difficult and time-consuming challenge.

Recent advances in deep learning methods have resulted in promising outcomes for a variety of computer vision problems. The growing concept in object recognition is toward deep learning approaches that outperform more traditional approaches for image classification. With the availability of large-scale datasets and convolutional neural networks (CNN), one of several deep learning techniques, remarkable progress has been made in image recognition [4]. In image recognition tasks, convolutional neural networks (outperform other state-of-the-art approaches. Classification of diagnostic images according to anatomy utilizing deep convolutional networks). Recently, CNN was investigated for potential use in medical image classification, and algorithms and models were suggested.

Roth et al. developed a two-tiered coarse-to-fine cascade architecture for a CAD device [5]. We generated regions of interest or volumes of interest and used them as feedback for a second tier. This second stage provided 2D (two-dimensional) or 2.5D views through scale transformations, random translations, and rotations. These random views are used to train classifiers using deep convolutional neural networks (CNNs). During research, the CNN assigns class probabilities to a new collection of random views, which are then combined to obtain a final classification chance for each candidate. Roth et al. proposed a CNN-based algorithm for classifying body parts in computed tomography (CT) photographs [6]. Patel et al. defined a system for classifying interstitial lung diseases using computed tomography (CT) images [3]. It used the whole picture as the input to a CNN. Shin et al. [4] explored three facets of computer-aided detection problems including deep convolutional neural networks. It examined two distinct CAD issues: the detection of thoraco-abdominal lymph nodes (LNs) and the classification of interstitial lung disease (ILD). Anthimopoulos et al. established a CNN for interstitial lung diseases (ILDs) classification [7]. The network is composed of five convolutional layers with 2. 2 kernels and Leaky ReLU activations, accompanied by three thick layers and average pooling with a scale equivalent to the final function map size. Khan et al. contrasted the output of three major deep learning architectures: LeNet, AlexNet, and GoogLeNet on medical imaging data comprising five anatomical structures for the purpose of anatomic basic classification [8]. Li et al. developed a Convolutional Neural Network (CNN) with a shallow convolution layer for the purpose of classifying lung picture patches associated with interstitial lung disease [9].

Most current approaches classify images using image patcher sub images, which complicates the task of patch region selection. The aim of our research is to develop an automated computer-aided classification system capable of classifying entire breast images without the need for feature extraction and selection. We suggested a customized convolutional neural network (CNN) for mammogram classification to address the above shortcomings. We propose a novel learning technique, reinforcement sample learning, to accelerate CNN model training due to the broad size of the input picture. Our customized CNN platform can automatically and easily learn the intrinsic picture characteristics from mammogram photographs that are most useful for classification. The rest of the paper is structured as follows: we introduce the proposed CNN model through a novel learning scheme. The final segment discusses the experimental findings and draw a conclusion.

## 2. Proposed Method

*2.1. Data preprocessing*

Prior to send the mammogram images to the CNN model for training, we perform several preprocessing measures, including breast region extraction and image size normalization. Apart from the breast, mammograms have additional regions such as a date mark and text marks. To isolate the breast region from the context, we use a thresholding-based approach to segment the breast region from the rest of the image, since the breast regions have a higher amplitude than the rest of the image. Then, using a mirror process, we turn the right breast to the image's left side. Finally, the representation of the breast area is standardized to a fixed size for use in the CNN model.

*2.2. CNN model construction*

We constructed the CNN model using four different forms of layers. There are four types of layers: convolution, pooling, fully linked, and loss feature. The CNN layers' comprehensive structure is listed in Table 1:

| Layer Number | Layer type | Filter Size |
|---|---|---|
| 1 | C | 5X5X32 |
| 2 | S | 3X3 |
| 3 | C | 5X5X32 |
| 4 | S | 3X3 |
| 5 | C | 5X5X64 |
| 6 | S | 2X2 |
| 7 | C | 5X5X64 |
| 8 | S | 3X3 |
| 9 | F | 64X64X2 |
| 10 | L | |

Table 1 Structure and configuration of CNN model (C: convolution layer; S: pooling layer; F: full connected layer; L: loss function layer)

*2.3. Reinforcement sample learning strategy*

The aim of reinforcement sample learning is to train the model on samples that performed poorly during the training procedure.

The following sections explain the procedure:

1. Divide the samples randomly into multiple batches.

2. Train the model using all sample batches in a single epoch.

3. Determine which batches are subpar relative to the current epoch.

4. Adjust the model by adding more epochs to these batches.

5. Proceed to phase 2 to train the model before it meets the termination criterion.

In Step 2, the criteria for identifying samples with suboptimal output is described as follows:

$$C = \frac{1}{N}\sum_{j=1}^{N} E_r(i)|$$

where $E_r(i)$ denotes the $i^{th}$ batch's classification error rate. The total number of batches is denoted by $N$.

If the batch's error rate exceeds the C, it will be trained for a larger number of epochs. The epoch number is calculated as follows:

$$MEp = f(Er(i) - C)$$

where **f (.)** denotes a function that converts the discrepancy between Er and C to the epoch number. In our experiment, we apply it using a piecewise function.

The complete algorithm can be outlined as follows:

Table 2 Detailed steps of the CAD classification algorithm

| |
|---|
| Input: Mammogram image |
| Output: Diagnosis result |
| Step 1: Using the thresholding approach and associated component analysis, determine the breast region. |
| Step 2: Crop and normalize the breast region of the original image. |
| Step 3: Build a 15-layer CNN. |
| Step 4: Train the CNN with normalized images using reinforcement sample learning scheme. |
| Step 5: Use a reinforcement sample learning scheme to train the CNN with normalized images. |

## 3. Proposed Method

### 3.1. Dataset

The experiment makes use of the Digital Database for Screening Mammography (DDSM) [10]. The DDSM is a mammographic image archive intended for use by researchers. The database's primary mission is to encourage sound research into the advancement of computer algorithms to assist in screening. The database can also be used for secondary purposes such as the production of algorithms to help in diagnosis and the creation of teaching or training aids. The archive contains 2103 mammograms of the left and right breasts performed in the Medio-Lateral Oblique (MLO) view. Both photographs are in gray-scale and come in a variety of sizes. Every case contains two photographs of each breast, as well as certain patient statistics, such as the patient's age at the time of the study, breast density level [11], and subtlety rating for anomalies. Suspicious places of images are correlated with pixel-level "earth reality" [12] knowledge regarding their positions and forms. 695 of the 2620 cases are classified as mild, 717 as benign [13], and 691 as malignant. In the current work, we include only the benign and malignant types while attempting to solve the classification issue.

### 3.2. Experiment Setup

Both experiments are conducted on a server equipped with two Intel Xeon six-core processors and 128GB of memory. Multiple NVIDIA Tesla K40 GPUs, each with 12GB of memory, are installed on the server. We use a single Tesla K40 GPU for our experiments, as the quick completion period of the requisite number of epochs eliminates the need for multiple GPUs. This is particularly valid given that the reinforcement sample learning method needs less epochs to achieve the same error as the conventional approach. Our experiment setup leaves us plenty of room to increase the complexity of the layers used in each epoch, since we can comfortably scale to several GPUs in the future.

### 3.3 Experimental results

We train the proposed CNN model using reinforcement sample testing on 1000 benign and 1000 malignant pictures. To confirm the training procedure, 450 benign and 450 malignant images are used. The classification error for the training and validation sets is illustrated in Figure 1b.

We compare the reinforcement sample learning procedure to the more conventional method of learning. Fig. 1 illustrates the error values after 150 epochs. According to the traditional strategy shown in Figure 1a, the top one error in the testing set is 11.5 percent at the conclusion of training, and the top one error in the validation set is also as big as 22.1 percent. Because of the enhanced technique shown in Figure 1b, the top one error in the testing set decreases to 11.5 percent at the conclusion of training, while the top one error in the validation set increases to 22.1 percent. By example, we can see that the proposed reinforcement sample training approach improves training speed and generalizes the model.

*3.4* Discussion

As shown by the preceding experiments and comparisons, the proposed reinforce sample learning approach will increase training speed while decreasing validation error. This is because the proposed method considers the weights or variables associated with various samples during the training process. For non-homogeneous samples with a large error rate, reinforcement learning needs further preparation and tuning.

We discovered the reason for the proposed CNN model's high error in the validation set after examining the findings and images:

1. The significant regions are a relatively limited portion of the whole image that provide important characteristics for invasive ductal carcinoma diagnosis.

2. Certain images are of low quality. Since the mammogram images in DDSM were acquired from film using a digital scanner, the quality of all digital images varies according to lighting circumstances, physical film images, and scanner setup and settings.

3. In DDSM, image variance is extremely high. Certain images in the validity set are wildly dissimilar to those in the training set.

4. The classification results in the labels refer to patients or cases rather than to specific images. It is probable that one breast has a tumor and the other does not in the image dataset.

In the future, we will continue to refine the CNN model to include the capability of identifying tiny important characteristics such as small masses and microcalifications. Additionally, we want to examine the images in DDSM and determine the correct label for each image. The newly refined CNN model will be conditioned and validated on images labeled with individual groups.

**4. Conclusion**

The aim of this study is to present an invasive ductal carcinoma classification algorithm that is focused on the novel use of convolutional neural networks on breast mammogram images to enhance feature extraction and training speed. The algorithm makes two contributions. To begin, a new CNN model is developed to automatically extract and classify features from mammograms with large image sizes. In comparison to standard classification methods, it is fully automated and eliminates the need for manually identifying regions of interest, detecting tumors and microcalcifications, and extracting and describing characteristics. Second, a novel reinforcement sample learning scheme is proposed to practice the CNN faster and with less iteration epochs. The suggested model is qualified and validated on the Digital Database for Mammography Screening (DDSM). A random sample of 1000 benign and malignant cases is used as the training set, whilst another 450 benign and malignant cases are used as the validation set. The experimental findings indicate that the CNN model with the latest training scheme will produce a 11.4 percent error within 150 epochs and a less than 22.1 percent error in the validation set. It shows that CNNs can be used to extract and classify features from vast medical images, and that deep learning can be used to train extremely efficient and precise classifiers that can be used to initialize more computer-aided diagnosis. Due to the proposed algorithm's fast training time, it will see widespread use in the image classification sector, due to the unique relationship between Convolutional Neural Networks and the Reinforcement Sample Learning Strategy used.